\documentclass[conference,a4paper]{APSIPA2020}
\usepackage{multirow}
\usepackage{url}
\usepackage{graphicx}
\usepackage{amsmath}
\usepackage[psamsfonts]{amssymb}
\usepackage{amsxtra}
\usepackage{threeparttable}

\newcommand\blfootnote[1]{%
  \begingroup
  \renewcommand\thefootnote{}\footnote{#1}%
  \addtocounter{footnote}{-1}%
  \endgroup
}

\graphicspath{{fig/}} 

\begin{document}
\title{Cost Sensitive Optimization of Deepfake Detector}

\author{%
\authorblockN{%
Ivan Kukanov*\authorrefmark{2}\authorrefmark{3}, Janne Karttunen*\authorrefmark{2}\authorrefmark{4}, Hannu Sillanp\"a\"a*\authorrefmark{2}\authorrefmark{4}, and Ville Hautam\"aki\authorrefmark{2}\authorrefmark{4}
% Ivan Kukanov\authorrefmark{1}, Kuan K. Teh\authorrefmark{2}, Helen Huong Thuy Ngoc Thai\authorrefmark{3},
% Dat H. Tran\authorrefmark{4} 
}

\authorblockA{
\authorrefmark{2}School of Computing, University of Eastern Finland, Finland\\ 
\authorrefmark{3}Institute for Infocomm Research, A*STAR, Singapore\\
E-mail: ivan@kukanov.com, \{jannkar, hannusi, villeh\}@uef.fi
% \\\normalsize{\{jannkar, hannusi, villeh\}@uef.fi, ivan@kukanov.com}
% Institute For Infocomm Research (I2R), A*STAR, Singapore \\
% E-mail: \{ivan\_kukanov,teh\_kah\_kuan,nthhthai,hdtran\}@i2r.a-star.edu.sg
}
%%
%\authorblockA{%
%\authorrefmark{2}
%Northwestern Polytechnical University, Xi'an, China\\
%E-mail: ynzhang@nwpu.edu.cn  Tel/Fax: +86-29-XXXXXXXX}
%
}

\maketitle
\thispagestyle{empty}

\begin{abstract}
  Since the invention of cinema, the manipulated videos have existed. But generating manipulated videos that can fool the viewer has been a time-consuming endeavor. With the dramatic improvements in the deep generative modeling, generating believable looking fake videos has become a reality. In the present work, we concentrate on the so-called deepfake videos, where the source face is swapped with the targets. We argue that deepfake detection task should be viewed as a screening task, where the user, such as the video streaming platform, will screen a large number of videos daily. It is clear then that only a small fraction of the uploaded videos are deepfakes, so the detection performance needs to be measured in a cost-sensitive way. Preferably, the model parameters also need to be estimated in the same way. This is precisely what we propose here.
\end{abstract}

\blfootnote{
* Equal contribution.\\
\authorrefmark{4} These authors were supported by the Academy of Finland (grant \#313970) and Finnish Scientific Advisory Board for Defence (MATINE) project \#2500M-0106. We gratefully acknowledge the support of NVIDIA Corporation with the donation of the Titan Xp and V GPUs used for this research. We thank Marko Vahela for collecting the deepfake videos.}

\section{Introduction}
\label{sec:intro}
%\textcolor{red}{
%Notes: \\
%- We need a nice "abstract figure". Let's plan what we %need for it. Some examples from our dataset? \\
%}

In a just few years, the attention of the general public and research community has been raised to the dangers of the deepfakes~\cite{stehouwer2019detection}. Deepfakes, in general, are defined as face swapped videos, where the {\em source} individual's face is swapped to the {\em target} individuals face. This is also known as an identity swap~\cite{stehouwer2019detection}. It is easy to imagine socially disruptive applications of such a technology~\cite{Chesney2019lawfare}, such as video of a politician in a questionable activity before elections. In addition, it has been shown that human observers can be fooled by the deepfakes~\cite{faceforensics++}. This raises a need to develop automatic methods for deepfake detection. Such methods could be then employed by streaming services, law enforcement personnel and individual citizens. 

Multiple ways exist in generating deepfakes~\cite{protectingLeaders}: Face-swap \cite{mesonet} swaps the face of a person with another person frame-by-frame, lip-sync methods modify mouth movements in the video to match a swapped sample of speech, puppet-master \cite{pagan} methods transfer movements from an actor to the target person. Generating swapped face images requires a high-quality generative model of face images. Such models are for example GAN models like StyleGAN \cite{stylegan}, FS-GAN \cite{fsgan} and the few-shot method in \cite{fewshotfakes}. The idea is that new face images are generated frame by frame. Then same face expressions and orientations of the target face would be automatically generated to the new face image. 

Being generated frame by frame, deepfakes can be detected based on cues such as inconsistent head poses \cite{headPoses} and eye blinking \cite{eyeBlinking}. Deepfakes can also be detected by training a deep classifier to focus on frame-by-frame artefacts   \cite{mesonet,celeb-df,capsulenetworks} and considering temporal differences between frames \cite{deepfakeLSTM}. As is expected, for known deepfake generation types low error rates are reported, but for the unseen attack type, collected from the Internet, the results are shown to be poor~\cite{celeb-df}. It is noteworthy that all previous studies consider equal costs for both miss classification rates (miss and false alarm). 

Some of the datasets are available for the development of deepfake detectors. Publicly available datasets are DeepfakeTIMIT \cite{deepfaketimit} and deepfakes subset from Faceforensics++ (FF++) \cite{faceforensics++} have been widely used for training and evaluation \cite{faceforensics++,celeb-df, multitaskJunichi, detectingWarpingArtifacts}. In multiple studies datasets have been collected directly from online sources \cite{mesonet, deepfakeLSTM}, utilizing videos created by regular users. Just recently more datasets have been emerging, such as newly added extension for FF++ dataset \footnote{\url{https://github.com/ondyari/FaceForensics/tree/master/dataset/\\DeepFakeDetection}} and Celeb-DF \cite{celeb-df}. 

We trained our models using the pooled  DeepfakeTIMIT and FF++. We also collected a number of deepfake videos made for entertainment purposes from the YouTube. This set works as a proxy for unseen deepfake condition. Both of these sets we released publicly. In addition, noting that it is expected that deepfakes are much more rarer than legitimate videos, we promote a cost sensitive measurement of deepfake detection performance. And finally, we finetune detection models to directly optimize the cost sensitive metric via {\em maximal figure-of-merit} (MFoM) framework~\cite{Gao:2006:MFL:1148020.1148022}.

\section{Detection Metrics}
\subsection{Detection Cost Function}
In this work, we use detection cost function (DCF) as the performance measure. It is the conventional performance measure in the speaker recognition domain for long time~\cite{Sadjadi2017}. It serves as a unified measure for evaluation a performance of detection models and gives insights on the new advanced methods. DCF is defined as a weighted sum of two types of errors: \textit{miss detection} $P_{\mathrm{miss}}$ and \textit{false alarm} (acceptance) $P_{\mathrm{fa}}$
\begin{align} \label{eq:c-dcf}
C_{\mathrm{DCF}} \left(t \right)  = C_{\mathrm{\mathrm{miss}}}  \cdot P_{\mathrm{tar}}  \cdot P_{\mathrm{miss}} \left( {t} \right) + \nonumber \\
+ C_{\mathrm{fa}}  \cdot \left( {1 - P_{\mathrm{tar}} } \right) \cdot P_{\mathrm{fa}} \left( {t} \right),
\end{align}
it depends on the decision threshold $t$, applied to the scores; parameters $C_{\mathrm{miss}}$ (cost of a miss detection) and $C_{\mathrm{fa}}$ (cost of a false alarm) are usually set to one; $P_{\mathrm{tar}}$ is \textit{a prior} probability of the target class, $P_{\mathrm{tar}}$ takes value from $\{0.1, 0.05, 0.01\}$. Empirical probabilities of miss detection and false alarm are
\begin{equation}
P_{\mathrm{miss}}(t)  = \frac{{FN\left( t \right)}}
{P} = \frac{{\sum\limits_{y_i  \in y_{\mathrm{tar}} } {\mathbf{1}\left( {g\left( {{\mathbf{X}}_i} \right) < t} \right)} }}
{P},
\end{equation}
and 
\begin{equation}
P_{\mathrm{fa}}(t)  = \frac{{FP\left( t \right)}}
{N} = \frac{{\sum\limits_{y_i  \in y_{\mathrm{non}} } {\mathbf{1}\left( {g\left( {{\mathbf{X}}_i } \right) \geq t} \right)} }}
{N},
\end{equation}
where the function $\mathbf{1} (\cdot)$ is the indicator function applied to the model scores $g(\textbf{X}_i)$ on every sample $\textbf{X}_i$, $P$ and $N$ are the total number of target and non-target samples.

\subsection{Equal Error Rate}
Another conventional detection performance measure is the equal error rate (EER).
The EER is expressed using the same $P_{\mathrm{miss}}$ and $P_{\mathrm{fa}}$, those are increasing and decreasing functions of the threshold $t$ and the value of EER is defined on the intersection. 
%In \cite{Poh2004Evidences}, authors present theoretical and empirical analysis of EER. 
The lower the value of EER the better the performance of a system. On the other hand, EER is defined as the equality
\begin{equation}\label{eq:eer_fnr_fpr_equality}
\mathrm{EER}(t^*) = P_{\mathrm{miss}}(t^*) = P_{\mathrm{fa}}(t^*),
\end{equation}
where an optimal threshold for the EER is $t^*$ and any threshold $t \in [0, 1]$. Criteria for the optimal threshold is
\begin{equation} \label{eq:eer_fnr-fpr}
t^* = \mathop {argmin}\limits_{t}|P_{\mathrm{miss}}(t) - P_{\mathrm{fa}}(t)|.
\end{equation}

%\begin{figure}
%\centering
%\includegraphics[width=7cm]{fnr_fpr_clean}
%\caption{\label{img:fnr_fpr_clean}
%	Empirical equal error rate (EER) is defined as the intersection of false negative and false positive rates.}
%\end{figure}

% In \cite{Poh2004Evidences}, they define ``theoretical'' FNR and FPR using Gaussian distribution and apply class-dependent variance reduction approach. Due to a finite number of data points, the ``empirical'' FNR and FPR are not smooth functions. Thus, we are not able to find the stable empirical EER, small changes of threshold $t$ causes a big changes in the EER \cite{Poh2004Evidences}.

\section{Maximal Figure-of-Merit Solution}
In this work we explore MFoM framework~\cite{Gao:2006:MFL:1148020.1148022} for DCF and EER optimization. The goal is to develop an objective function which directly optimizes the measures of performance.
\subsection{Discriminant Function} 
The inference of MFoM framework begins with the definition of the \textit{discriminant function}. For a neural network, it is the activation scores $g(\textbf{X}| \mathbb{W})$ of the output layer, which defines the confidence of a model on a particular input sample $\textbf{X}$. The choice of a proper discriminant function depends on the nature of the classifier, and the task at hand. Discriminant functions are defined on the classifier parameters set $\mathbb{W}$. 

% The goal is to find the optimal set of parameters that  minimizes the objective function and the discriminant functions must satisfy the decision rule for any sample $\textbf{X}_i$ of class $C_k$
% \begin{equation} \label{eq:decision_rule_discr_fun}
% g_k \left( {{\bf{X}}_i ; \mathbb{W}} \right) > g_j \left( {{\bf{X}}_i; \mathbb{W}} \right),
% \end{equation}
% where $k \in \textbf{y}_{\left\{ 1 \right\}}$ is the set of indices corresponding to units from a label vector $\textbf{y}$, accordingly $j \in \textbf{y}_{\left\{ 0 \right\}}$ is the set of indices corresponding to zeros from a label vector $\textbf{y}$. We notice here that the condition in (\ref{eq:decision_rule_discr_fun}) have unique $k$ for any sample $\textbf{X}_i$, in case of \textit{single-label} classification, because $\textbf{X}_i$ belongs to a single class $C_k$. 

\subsection{Misclassification Measure}

The next part of MFoM is a \textit{misclassification measure} \cite{Bishop2006}. This approach, based on misclassification measures, allows us to define different strategies for decision rules based on discriminant scores. In the previous studies for phonetic feature detection \cite{Kukanov2016Deep, Kukanov2020Maximal} and acoustic events detection \cite{KukanovMaximal}, authors proposed the \textit{units-vs-zeros} misclassification measure for each class $C_k$ as

\begin{equation} \label{eq:misclas_measure_uvz}
 \psi_k = - g_{k} +   
 \frac{1}{\eta }\ln \left( {\frac{1}{{\left| {\bf{I}} \right|}}\sum\limits_{j \in {\bf{I}}} {e^{\eta g_j} } } \right),
	\end{equation}	
	
	\begin{equation}
\left\{ \begin{array}{l}
\text{if} \,\,\, C_k \,\,\,\text{is}\,\,\,1 \Rightarrow {\bf{I}} = {\bf{y}}_{\left\{ 0 \right\}},  \\ 
\text{if} \,\,\, C_k \,\,\,\text{is}\,\,\,0 \Rightarrow {\bf{I}} = {\bf{y}}_{\left\{ 1 \right\}},  \\ 
 \end{array} \right.
	\end{equation}	
where $\psi_k$ is defined for current sample $\bf{X}$ and its label $\bf{y}$; $\bf{I}$ is an index set, ${\bf{y}}_{\left\{ 1 \right\}}$ is the set of unit indexes and ${\bf{y}}_{\left\{ 0 \right\}}$ is the set of zero indexes in the label vector $\bf{y}$; $g_k$ are the discriminant functions; $\eta$ is a positive real-valued smoothing constant ($\eta = 1$ in our experiments). 

The first term on the left-side of (\ref{eq:misclas_measure_uvz}) is called the \textit{target model} and the right-side is the \textit{Kolmogorov mean} (generalised $f$-mean) \cite{Tikhomirov1991Notion} of the competing (\textit{confusing}) models. The misclassification measure is the differences between the target class and the average of the confusing classes. 

The sign of the misclassification measure indicates the correctness of classification: $\psi_k(\cdot) \leq 0$ indicates the predicted class is correct, and $\psi_k(\cdot) > 0$ implies incorrect decision. The absolute value of the $\psi_k$ quantifies the separation between the correct and competing classes \cite{KatagiriNew}. The equality $\psi_k(\cdot) = 0$ defines the decision boundary between a class $k$ and the rest.

\subsection{Smooth Error Counter}

The third block of the MFoM framework is the \textit{smooth error counter}, which plays the key role for the approximation of discrete performance measures based on discrete error counts (i.e., false positive and false negative statistics)
% We therefore introduce a smooth (differentiable), and monotonic approximation function that squeezes the output of the \textit{misclassification measure} to the $[0, 1]$ range. That squeezing function can be a sigmoid, a hinge, an exponential, or any other smooth function approximator. In this paper, the sigmoid function is used to approximate the discrete error count of misclassified samples; it is a smoothed version of the error step function
\begin{equation} \label{eq:loss_func}
l_k = \frac{1}{{1 + \exp \left[ { - \alpha_k \psi_k - \beta_k} \right]}},
\end{equation}
where $k = \overline{1, M}$ is the class index, $\alpha_k$ and $\beta_k$ are real valued parameters of the scale and shift transformation. From deep learning point of view, we can interpret the linear transformation ($\alpha_k$ and $\beta_k$) of the \textit{misclassification measure} as an additional layer of a network. In this work, we propose the optimization of those parameters similar to the batch normalization technique, when the error of the objective function $E$ is backpropagated through $\alpha_k$ and $\beta_k$  as well
\begin{equation}
\frac{{\partial E}}{{\partial \alpha _k }} =  - \frac{{\partial E}}{{\partial l_k }} \cdot \psi _k,
\end{equation}
\begin{equation}
\frac{{\partial E}}{{\partial \beta _k }} =  - \frac{{\partial E}}{{\partial l_k }}.
\end{equation}

\subsection{Approximation of DCF Objective}

The key ingredients of the proposed MFoM framework are: a) discriminant function, which in our case are output scores of a network model, b) misclassification measure (\ref{eq:misclas_measure_uvz}), and c) smoothed error counter (\ref{eq:loss_func}). Now that these components have been introduced, we can express the DCF in terms of those three entities within the deep neural network paradigm.	We introduce a smooth approximation of discrete error rates $P_\mathrm{miss}(t)$ and $P_\mathrm{fa}(t)$, for this purpose we apply the smooth error counter from (\ref{eq:loss_func})

\begin{equation}\label{eq:batch_FN_approx}
\hat{P}_\mathrm{miss} \buildrel \Delta \over = \frac{\sum\limits_{k = 1}^M {FN_k }}{P} = \frac{\sum\limits_{k = 1}^M {\sum\limits_{{\bf{X}} \in \mathbb{T}} {l_k \cdot {y_k}} }}{P} ,
\end{equation}
\begin{equation}\label{eq:batch_FP_approx}
\hat{P}_\mathrm{fa} \buildrel \Delta \over = \frac{\sum\limits_{k = 1}^M {FP_k }}{N}  = \frac{\sum\limits_{k = 1}^M {\sum\limits_{{\bf{X}} \in \mathbb{T}} {\left( {1 - l_k} \right) \cdot {\overline{y}_k}}}}{N},
\end{equation}
where $y_k$ and $\overline{y}_k$ are the binary labels and their inverse, assigning sample ${\bf{X}}$ to class $k$. Eventually, the MFoM-DCF objective function is obtained  and applied for DNN parameters (${\mathbb{W}}$), optimized on a training set ${\mathbb{T}}$
\begin{align}\label{eq:mfom_dcf_obj}
 & E_{\mathrm{DCF}} (\mathbb{W} | \mathbb{T})  = P_{\mathrm{tar}}  \cdot \hat{P}_{\mathrm{miss}} \left( \mathbb{W} | \mathbb{T}\right)  + \nonumber \\ 
&\qquad{}  +  \left( {1 - P_\mathrm{tar} } \right) \cdot \hat{P}_{\mathrm{fa}} \left( \mathbb{W} |\mathbb{T} \right),
 \end{align}
where $P_\mathrm{tar}$ is \textit{a prior} probability from (\ref{eq:c-dcf}). 

\subsection{Approximation of EER Objective}
Similar to the DCF approximation using the MFoM framework, we can embed the EER into the objective function for DNN optimization. Using two properties of the discrete EER (\ref{eq:eer_fnr_fpr_equality}) and (\ref{eq:eer_fnr-fpr}), we infer smoothed MFoM-EER objective, which is transformed to unconstrained function
\begin{align}\label{eq:mfom_eer_obj}
& E_{\text{EER}} \left({\mathbb{W}|\mathbb{T}} \right) = \hat{P}_{\mathrm{\mathrm{fa}}} \left( {\mathbb{W}|\mathbb{T}} \right)+ \nonumber \\ 
&\qquad{} + \lambda \left| {\hat{P}_{\mathrm{miss}} \left( {\mathbb{W}|\mathbb{T}} \right) - \hat{P}_{\mathrm{fa}} \left( {\mathbb{W}|\mathbb{T}} \right)} \right|,
\end{align}	
where $\lambda \ne 0$ is a Lagrange multiplier, in the experiments we assign $\lambda = 0.5$.

\section{EXPERIMENTS}
\label{sec:experiments}

\subsection{Detection methods}
\label{sec:DetectionMethods}

For the baseline methods we use CNN-based approach and recurrent {\em long short-term memory} (LSTM) network. While CNN implementations for deepfake detection exist \cite{mesonet}, we chose to train one from scratch, using the ready implementation of MobileNet \cite{mobilenet}. For the LSTM, we implemented a network based on the description in \cite{deepfakeLSTM}, using similarly pretrained InceptionV3 for feature extraction, and then LSTM for the temporal analysis. Subsequences of 20 frames were used as an input for network. So, network produced score after 20 frames. For a single video with length more than 20 frames, subsequence scores were averaged.  

\subsection{Dataset}
\label{sec:Dataset}
For training data we have used the FaceForensics++ and Deepfake-TIMIT datasets merged together. FaceForensics++ includes data for both real and fake classes. For Deepfake-TIMIT, corresponding pristine videos from VidTIMIT \cite{vidtimit} are used for data in the real class. Deepfake-TIMIT includes face swaps made using a higher and lower quality model, of which both are included in our merged data. Two versions of all videos were included with different values for the Constant Rate Factor (CRF) compression parameter: high quality (CRF=23) and low quality (CRF=40). Facial images were cropped and aligned from the video frames at resolution 256x256 using DeepFaceLab software\footnote{\url{https://github.com/iperov/DeepFaceLab}}%, which uses work from \cite{s3fd} for detection and \cite{face_alignment} for alignment. The results of the automatic face extraction were manually inspected to remove misdetections. The dataset was split into subsets for training, validation and testing. The latest addition in FaceForensics++, the Deep Fake Detection Dataset from Google and JigSaw, was not available early enough to be used in this work.

%. with parameters $\alpha = 1, \rho = 1$ and resolution 256x256. \todo{other hyperparameters?}

To additionally evaluate our methods in a scenario more accurate to real-life, we have collected a dataset from YouTube including deepfakes generated by regular users. These deepfakes were originally created for entertainment purposes, which is why they are more fine-tuned and polished than our initial training set. The generation algorithms of these deepfakes are unknown, and unseen to our methods, which is why the detection rate is expected to be lower, similarly as in previous studies \cite{celeb-df}.

We manually downloaded and annotated 79 deepfake videos from YouTube, extracting total of 98 face swaps. For the real data, we similarly downloaded 98 samples annotated in VoxCeleb2 dataset \cite{voxCeleb2}. All the videos were further divided to scenes, to make temporal analysis possible. Our collected dataset will be available for download\footnote{\url{http://cs.uef.fi/deepfake\_dataset/}}.

\subsection{Results}
\label{sec:Results}

% Performance of the fine-tuned models with MFoM-base objective functions on the test dataset. Models are trained with the categorical cross-entropy objective (CE baseline) and tuned with MFoM-EER or MFoM-DCF objectives.

\begin{table}[htb]
\caption{\it Performance results on test dataset and self collected evaluation dataset, denoted by Eval. MEER and MDCF signifies MFoM with soft EER and DCF objectives. Results are shown in EER (\%) and DCF with corresponding $P_{\mathrm{tar}}$.}
\label{tab:detection_scores}
\vspace{5mm}
\centerline{
\begin{tabular}{|c|c|ccc||c|}
    \hline
    \multirow{2}{*}{Method} & \multirow{2}{*}{EER} & \multicolumn{3}{c||}{minDCF with $P_{\mathrm{tar}}$} & Eval \\
    &  & 0.1 & 0.05 & 0.01 & EER \\
    \hline \hline
    LSTM \cite{deepfakeLSTM} & 24.1 & 0.88 & 0.92 & 0.96 & 38.90\\
    %\cline{2-3}
    CNN & 8.07 & 0.37 & 0.43 & 0.60 & 30.80\\
    \hline
    %\cline{2-3}
    %CNN+Attention & 6.55 & \textbf{0.26} & \textbf{0.32} & \textbf{0.43} & 32.90\\
    CNN+MEER & 7.16 & 0.44 & 0.50 & 1.00 & 32.32\\
    CNN+MDCF\_0.1 & \textbf{6.03} & 0.33 & 0.46 & 0.88 & 32.49\\
    CNN+MDCF\_0.05 & 6.67 & \textbf{0.28} & 0.32 & 0.46 & \textbf{30.56}\\
    CNN+MDCF\_0.01 & 6.72 & 0.35 & 0.43 & 0.56 & 32.09\\
    \hline
\end{tabular}}
\end{table}

\begin{figure}[!htb] 
\begin{minipage}[b]{\linewidth} 
  \centering
  \centerline{\includegraphics[width=6.0cm]{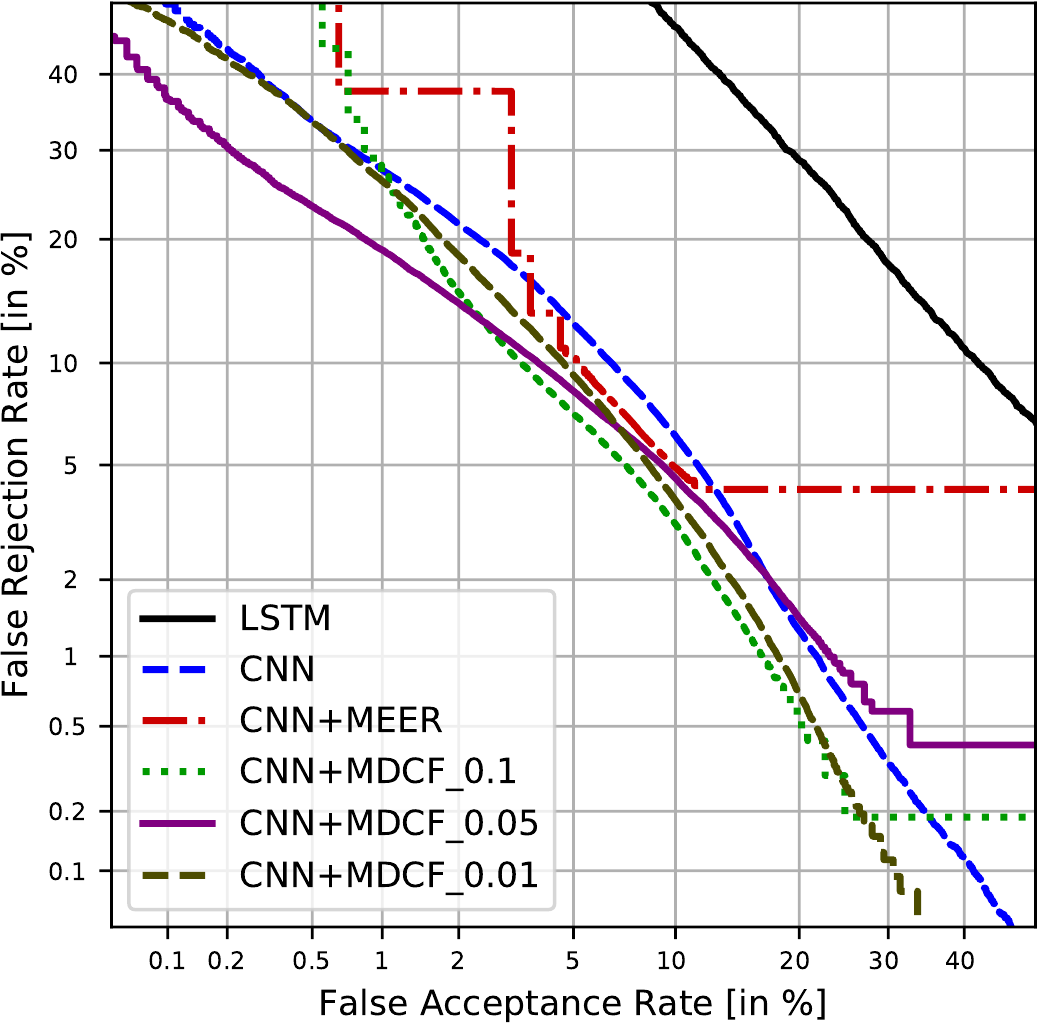}}
%  \vspace{1.5cm}
  \centerline{(a)}\medskip
\end{minipage}
\hfill
\begin{minipage}[b]{\linewidth}
  \centering
  \centerline{\includegraphics[width=6.0cm]{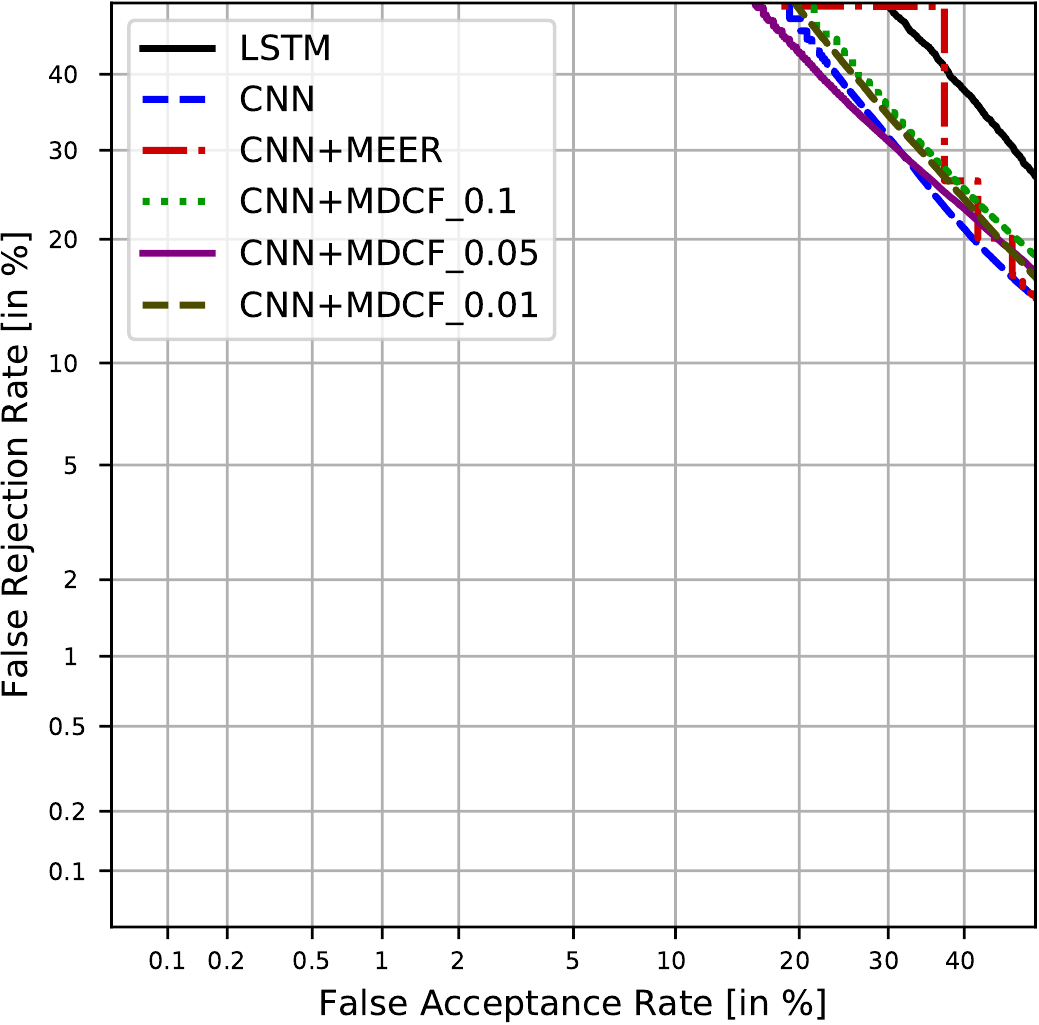}}
%  \vspace{1.5cm}
  \centerline{(b)}\medskip
\end{minipage}
\caption{Detection error tradeoff (DET) curves on the test (a) and evaluation sets (b) for the evaluated deepfake detectors. }
\label{fig:det_test_eval}
\end{figure}

The results of detection methods are shown in Table~\ref{tab:detection_scores}. We did not reach similar accuracy with LSTM as reported in \cite{deepfakeLSTM}, which can be explained by the different dataset or implementation differences. With the CNN, we obtained 8,07\% EER, %, which was further improved to 6.55\% EER using the attentive pooling. 
%We experimented with the baseline CNN detector, applying ,
on which we applied MFoM-based objectives for the further fine-tuning. We performed fine-tuning for 5 epochs with MFoM-EER and MFoM-DCF, i.e. MEER, MDCF\_0.1, MDCF\_0.05 and MDCF\_0.01, respectively. Results are shown in Table~\ref{tab:detection_scores}. Consistent improvement was obtained. The best performance in terms of EER, was obtained   by optimizing CNN with MFoM-DCF and prior of 0.1 (CNN+MDCF\_0.1). %However, in terms of the minDCF the best model is CNN with attention. This is because MFoM-based objectives mostly optimized EER decision point, as shown in Fig.~\ref{fig:det_test_eval}.
These methods are additionally evaluated on the collected dataset (Table~\ref{tab:detection_scores}, rightmost column). As expected, the results were significantly worse due to better deepfake quality. The best detection method was CNN with MFoM-DCF tuning (CNN+MDCF\_0.05), it scored only 30.56\% EER. Analysis of results showed that videos with lower deepfake quality were classified correctly as deepfakes. However, most of the deepfakes in evaluation set surpassed our training set in terms of quality, which is why tampered videos were not caught with our detection methods.

\section{Conclusions}
In this work, we defined the deepfake detection task as a cost-sensitive objective. We borrowed the measurement technology from the NIST SRE campaigns. The idea being that essentially both tasks, NIST SRE and deepfake detection, are screening tasks. We then defined a cost-sensitive optimization technique, utilizing the MFoM theory. We showed that fine tuning the CNN model with MFoM improves the EER from the 8.07\% to 6.03\%. %In addition, we applied attentive pooling to CNN model and received improvement in EER from 8.07\% to 6.55\%. 

In our self collected evaluation set, we noticed that the best test set EER of 6.03\% is increased to more than 30\%. 
%Noting that estimating the detection performance on the unseen attack scenario, unseen deepfake generator, we collected our own evaluation set.
%This result is in line with previously published studies on the unseen attack conditions. 
The set is based on the deepfakes generated for entertainment purposes and uploaded to YouTube. We take these deepfakes to be a worst cases a user of the detector would encounter in practice. As a future work, we plan to investigate deepfake detector models that perform more robustly on the unseen conditions. In addition, we will expand the use of MFoM technique to our attentive pooling models, with the hope of further improvement on the detector performance. 

% \section*{Acknowledgment}

\bibliographystyle{IEEEtran}
% \bibliography{short_lib}
% \bibliographystyle{IEEEbib}
\bibliography{strings,refs}
\end{document}